\documentclass{article}

\usepackage{arxiv}

\usepackage[utf8]{inputenc} 
\usepackage[T1]{fontenc}    
\usepackage{hyperref}       
\usepackage{url}            
\usepackage{booktabs}       
\usepackage{amsfonts}       
\usepackage{nicefrac}       
\usepackage{microtype}      
\usepackage{cleveref}       
\usepackage{lipsum}         
\usepackage{graphicx}
\usepackage{natbib}
\usepackage{doi}

\usepackage{algorithm}
\usepackage{comment}
\usepackage{algorithm,algpseudocode}

\title{Adaptive Prototypical Networks}

\author{ 
    Manas Gogoi\\
	Department of Information Technology\\
	Indian Institute of Information Technology\\
	Allahabad \\
	\texttt{pcl2017001@iiita.ac.in} \\
	\And
	Sambhavi Tiwari \\
	Department of Information Technology\\
	Indian Institute of Information Technology\\
	Allahabad \\
	\texttt{rsi2018503@iiita.ac.in} \\
    \And
	Shekhar Verma \\
	Department of Information Technology\\
	Indian Institute of Information Technology\\
	Allahabad \\
	\texttt{sverma@iiita.ac.in} \\
}

\renewcommand{\shorttitle}{\textit{arXiv} Template}

\hypersetup{
pdftitle={Adaptive Prototypical Networks},
pdfsubject={q-bio.NC, q-bio.QM},
pdfauthor={David S.~Hippocampus, Elias D.~Striatum},
pdfkeywords={Prototypical Networks, Adaptive, Meta Learning},
}

\begin{document}
\maketitle

\begin{abstract}
	 Prototypical network for Few shot learning tries to learn an embedding function in the encoder that embeds images with similar features close to one another in the embedding space. However, in this process, the support set samples for a task are embedded independently of one other, and hence, the inter-class closeness is not taken into account. Thus, in the presence of similar-looking classes in a task, the embeddings will tend to be close to each other in the embedding space and even possibly overlap in some regions, which is not desirable for classification. In this paper, we propose an approach that intuitively pushes the embeddings of each of the classes away from the others in the meta-testing phase, thereby grouping them closely based on the distinct class labels rather than only the similarity of spatial features. This is achieved by training the encoder network for classification using the support set samples and labels of the new task. Extensive experiments conducted on benchmark data sets show improvements in meta-testing accuracy when compared with Prototypical Networks and also other standard few-shot learning models.
\end{abstract}


\section{Introduction}
Supervised Learning with deep networks, although performs very well in domains with large amounts of data, it performs rather poorly in domains where the data distribution is scarce or has a long tail. With the aim of few-shot learning, this issue is mitigated quite convincingly by meta-learning algorithms. One such meta-learning algorithm is Prototypical networks \cite{snell2017prototypical}. The working of Prototypical networks, like other metric-based meta-learning algorithms, is based on learning an embedding function or an embedding space such that a non-parametric learner can easily classify the images in the learned embedding space. In particular, prototypical networks learn an embedding function with euclidean distance as a similarity metric between the query embedding and the class prototype, which is the mean vector of the support set embeddings. This results in similar-looking images getting embedded close to each other in the embedding space and forming clusters.

However, along with clustering of similar-looking images, it is more or less equally desirable to have separation of embeddings based on the distinct class labels when it comes to the task of classification. In prototypical networks, each image in a task is embedded independently, and hence there is no information shared between the embeddings regarding the similarity of classes. Thus, in the presence of similar classes or classes with similar features (for, e.g. the English alphabet 'p' and Greek alphabet '$\rho$') in a particular task, the embeddings from the similar classes lie very close to one another and also may overlap in some region. Subsequently, the class prototypes of similar classes in a task also end up forming very close to one another. Due to this small margin of separation between the prototypes, classifying the embeddings of the query samples becomes difficult. Moreover, the embeddings of the query samples may also get embedded in an overlapping manner due to the same reasons. This is especially a problem when evaluating the model on a new task during meta-testing, as the model does not have a mechanism to adapt to the learning error, unlike the meta-training phase.

In this paper, we describe an approach wherein, at the time of meta-testing, the support set of the new task is used to obtain the necessary knowledge to segregate the embeddings based on the distinct class labels. This is done by using the encoder of the model for classification using the support set and using the fine-tuned weights to re-embed all the data samples of the entire task. This acts as a mechanism to push the embeddings and, consequently, the prototypes away from each other, even in the presence of similar-looking classes. The proposed approach has been evaluated on several benchmark datasets, such as Omniglot and MiniImagenet, and comparisons have been made with existing state-of-the-art few-shot learning algorithms with different task configurations. The approach is simplistic and is yet able to improve classification accuracy under the circumstances of similar classes being considered.

\section{Preliminaries}
\label{sec:headings}

\subsection{Learning to Learn}
Learning how to learn is the notion of learning how to use the experience gained from performing tasks in the past in order to learn a new task quickly or more efficiently. This is also known as meta-learning.  Meta-learning is concerned with two levels of learning: the lower level rapidly learns how to perform a base task such as classification on a dataset; the higher level gradually learns the structure of the tasks in general \cite{santoro2016meta}. The higher level learning happens in due course of performing many such base tasks, and it also aids in the lower level learning process. The parameters (if any) that are learned in the lower level learning are called the task-specific parameters as they perform any one base task, while the parameters learned in the higher level learning are called the meta parameters as they capture a meta-knowledge about all the tasks in general. 
The meta-learning approaches can be widely classified into three broad classes: non-parametric, blackbox and optimization based. The non-parametric approaches \cite{koch2015siamese}, \cite{vinyals2016matching}, \cite{sung2018learning} are based on learning an embedding space where a non-parametric model can perform classifications with only a few shots in a task. The blackbox methods \cite{santoro2016meta} , \cite{munkhdalai2017meta}, \cite{mishra2017simple}, \cite{garnelo2018conditional} work by training the model or changing the internal state to specifically output the task-specific parameters. The optimization-based approaches \cite{finn2017model}, \cite{nichol2018first}, \cite{andrychowicz2016learning},  work by solving a bi-level optimization problem, where an inner loop optimizes the model for performing a base task and an outer loop optimizes the model to learn more general features of the task distribution.\\

The idea of meta-learning has also been adopted and widely used in the domain of reinforcement learning\cite{wang2016learning},\cite{yu2020meta},\cite{co2021evolving},\cite{finn2017model}, where an agent learns new tasks via only a few trajectories and rewards.
Also, the use of meta learning spans across different application areas such as few shot image detection\cite{fan2020few}\cite{rajeswaran2019meta}, facial recognition\cite{zhang2022meta}\cite{zeng2022face2exp},  robotics\cite{dasari2019robonet}\cite{song2020rapidly} etc.

\subsection{Episodic Learning}
The idea of episodic learning\cite{laenen2021episodes} is a gold standard for few-shot learning and meta-learning models in terms of training as well as performance evaluation. Considering a classification problem, a K-way N-shot learning configuration refers to a setting where the few-shot learning model is provided with N samples of each class to learn from and classify a test sample into one of K possible classes. A setting as such is called a task.

Few shot learning models involve generating sets of tasks by sampling data from a very large dataset. Each task has a support set ($D^{tr}$) and a query set ($D^{ts}$), and these are combinedly called an episode. In the case of classification, an episode is generated by randomly sampling a subset of labels, sampling examples for each of these labels and dividing the examples into a support and a query set. The support set consists of the labeled examples, $D^{tr} = {(x_{1}, y_{1}), (x_{2}, y_{2}), .... (x_{N}, y_{N})}$. The query set consists of the examples, $D^{ts} = {(x_{1}, y_{1}), (x_{2}, y_{2}), .... (x_{q}, y_{q})}$, which may or may not be labelled, depending on whether we are training or evaluating. In the context of meta-learning, the set of episodes used for training the meta-learning model is called the meta-train set and the set of episodes used for evaluation is called the meta-test set.

\subsection{Prototypical Networks}
Prototypical Networks \cite{snell2017prototypical} is a metric-based meta-learning model that works on the principle of finding an embedding space in which points lie in clusters around a class representative or prototype for all tasks. The construction of prototypical networks model consists of an encoder network ($f$) and a distance-based non-parametric classifier. The encoder network is a convolutional neural network that produces embeddings of the images in the task data-set. The mean vector of the support set embeddings are then computed for each of the classes in the task and these mean vectors are called the class prototypes.
\[r_{k}=\frac{1}{|D^{tr}|}\sum_{x_{i},y_{i}\in D^{tr}}f(x_{i})\], where $r_{k}$ denotes the prototype of a class label k.

The prototypes are then used by the non-parametric classifier to classify the query set embeddings. The nonparametric learner employs a distance function ($d$) that is used to compute the Euclidean distance between a query embedding and each of the class prototypes and output a soft distribution of the class predictions.
\[p_{\phi}(y=k|x)=\frac{exp(-d(f_{\phi}(x), r_{k}))}{\sum_{k'\in K}exp(-d(f_{\phi}(x), r_{k'}))} \], where K is the set of all class labels in a K-way classification task. Thus, the class label of the prototype that is closest to the query embedding is assigned the highest probability.

During the meta-training phase, the true labels of the query set and the predicted labels from the non-parametric classifier are used to compute the loss. In other words, the loss is the non-negative likelihood of the predicted class distribution  $-log \hspace{2pt} p_{\phi}(y=k|x)$, which is minimized in the training process of the encoder.

Performance evaluation on a new task during meta-testing is done by finding the class prototypes from the support set embeddings and predicting the class output of the query embeddings using the distance function and a hard classification. The true labels of the query are then compared with respect to the predicted class output for accuracy.

\begin{figure*}[h]
     \centering
     \includegraphics[width=0.8\textwidth, clip, keepaspectratio]{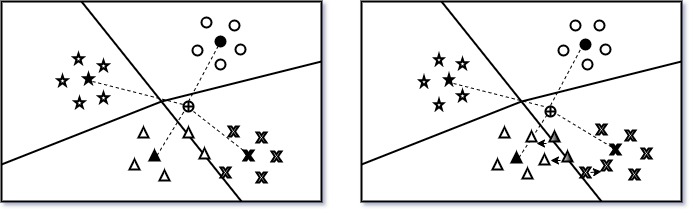}
     \caption{Schematic diagram of the embedding space before and after the proposed approach. \textbf{Left :} In the presence of classes with similar features, the support set embeddings (shown by different shapes) may overlap and prototypes (shown by filled shapes) may lie in close proximity. \textbf{Right :} After the proposed approach, the embeddings of one class get pushed away from other classes (shown with dark arrows) and forms cluster to give clearly separated prototypes. The query set embedding (shown by ex-or shape) gets clearly classified (to the class shown by 'cross' ) based on the Euclidean distance (dashed line segment). }
     \label{fig:iproto}
\end{figure*}

Prototypical networks provide a formulation for both few-shot as well as zero-shot learning scenarios. This is illustrated in Fig \ref{fig:iproto}.

\section{Related Work}
\label{sec:others}
There have been numerous works in the direction of meta-learning approaches.
In meta-learning, the idea of fine-tuning the model for classification is principally used in optimization-based approaches such as MAML\cite{finn2017model}, FOMAML\cite{nichol2018first}, ANIL \cite{raghu2019rapid}, Reptile \cite{nichol2018first}, iMAML. But the working principle of these approaches is entirely different, i.e. to obtain a suitable initialization in the meta-training process that enables rapid learning within a task.

There have also been a number of works that specifically aimed to improve upon Prototypical networks. Infinite Mixture Prototypes \cite{allen2019infinite} improve upon prototypical networks in terms of robustness with the formation of more than one cluster for a single class. A new cluster is formed with the support embedding as the cluster mean if the embedding is further away from all the cluster means than a threshold distance, $\lambda$. Here, however, the issue mainly addressed is different, i.e. to enhance performance on tasks such as alphabet recognition, which assume a multi-modal data distribution within the same class and little or no improvement in unimodal tasks.

One work that closely resembles our objective is matching networks \cite{vinyals2016matching}, which encourages embedding the images in the context of the entire support set. It discusses the concept of full context embedding, where a bi-directional LSTM is used to embed a support image at every time step. However, this induces an unnecessary dependence on the positional order in which the images are inputted. This is not the case in our approach, and the order of images is immaterial.

Further Meta prototypical Networks \cite{wang2021meta} also tackles a similar problem such that it introduces an inner loop of training in the model that involves training with a Euclidean distance-based loss for intraclass embeddings to be close and interclass embeddings to be far. However, this requires the model to be meta-trained entirely with both loops of optimization and provides good results when evaluated on a dataset other than the meta-training set. On the other hand, our approach requires training the model only during meta-testing; this enables the reuse of a pre-trained prototypical network and eliminates the need to retrain with the entire meta-training data.

\section{Methodology}

The idea of prototypical networks is to train an encoder ($f$) such that the embeddings of images from the same class lie in the proximity of one another, which makes it easy for a non-parametric learner to classify the points in a few shots. But since the embeddings are done independent of other images from the episode, the encoder actually embeds similar images or images with similar features close to each other in the embedding space irrespective of the class label. To tackle this problem, we introduce an approach wherein the encoder embeds each image by taking all the classes of the episode into consideration. \\

In our approach, the meta-training phase is conducted in the same way as is done in vanilla prototypical networks. However, in the meta-testing phase, the encoder network ($f$) is augmented to additionally attach a linear classifier layer at the end that transforms the encoder to an augmented encoder classifier ($F$). From the support set, an image is fed into this augmented encoder classifier ($F$), for which it computes the embeddings and classifies it into one of the K classes. At this stage, the embeddings are still computed independently. However, after all the support set images are classified in a similar fashion, the cumulative cross-entropy loss or error ($\mathcal{L}_{CE}$) is derived after comparison with the true labels of the support set images.
\[Loss_{clf}=\frac{1}{|D^{tr}|}\sum_{x_{i},y_{i}\in D^{tr}}\mathcal{L}_{CE}\{F(\theta_{enc},\theta_{cl}, x_{i}), y_{i}\}\], where $\theta_{enc}$ and $\theta_{cl}$ are the encoder and classifier layer weights respectively.

This loss is back-propagated and is used to train the encoder network weights ($\theta_{enc}$) along with the classifier layer weights ($\theta_{cl}$). The classifier layer is then detached from the encoder, and an updated encoder ($f'$) with weights ($\theta^{'}_{enc}$) is obtained.
\[\theta_{enc}^{'} \leftarrow \theta_{enc} - \nabla_{\theta_{enc}}Loss_{clf}\]
This updated encoder is used for further re-embedding the support and query set images to be fed into the non-parametric classifier.

This approach is novel to metric-based meta-learning models. Traditionally, metric-based models employ the support set only for non-parametric learning within a task and in reality, do not utilize the meta-testing support set for updating the parametric model. However, in this approach, the meta-testing support set is used for training the augmented encoder classifier for classification. The after-effects of training the encoder for classification is that when the images are re-embedded with the updated encoder, the embeddings of one class are intuitively pushed away from the others. This is because, during training, the augmented encoder classifier weights are optimized to distinguish the support set samples. Consequently, the encoder weights are necessarily updated to accommodate the maximal distinction in the embeddings before feeding into the final classifier layer for classification.

As a result, even in the presence of similar-looking classes, the re-embeddings of images done with the updated encoder tend to be separated based on the distinct class labels. Simultaneously, support images from the same class are re-embedded closer to each other to form compact clusters after training using the support set. Most importantly, this further causes the class-respective prototypes to form distinctively apart from one another.

Eventually, when the query samples are embedded using the updated encoder, the embeddings are formed close to the re-embedded support set clusters. Thus, the query embeddings are unambiguously classified according to the distance from the newly formed, clearly separated class prototypes.

The algorithm for the proposed approach is given in Algo \ref{alg:algo}

\begin{algorithm}[h]
   \caption{Proposed Approach}
   \label{alg:algo}
\begin{algorithmic}[1]
   \State {\bfseries Input:} (a) Meta Testing tasks, $D_{i} = {(data^{tr}_{i}, data^{ts}_{i})}$, for $i= 1,2,3....n$ (b) meta trained encoder weights $\theta_{enc}$
   \State Initialize the encoder f with $\theta_{enc}$.
   \State Augment f by adding classifier layer with weights ($\theta_{cl}$) at the end to make encoder-classifier F 
   \For{$task_{i}$ in $tasks$}
   \State $Loss_{clf}$ $\leftarrow$ $F(\theta_{enc}, \theta_{cl}, data^{tr})$
   \State $\theta_{enc}^{'}$ $\leftarrow$ $\theta_{enc}$ - $\nabla_{\theta_{enc}}Loss_{clf}$
   \State $p \leftarrow find\_prototypes$ ($\theta_{enc}^{'}$, $data^{tr})$
   \State $accuracy$ $+=$ $compute\_accuracy(p, data^{ts}) $
   \EndFor
\end{algorithmic}
\end{algorithm}

\section{Results and Discussion}
In this section, we describe the experimental details and results achieved with our proposed methodology and make comparisons with the original prototypical networks approach and other state-of-the-art meta-learning algorithms. Initially, the model is meta-trained with around 60000 batches of tasks sampled from a particular dataset, and the parameters are saved. From then on, the same pre-trained model is used for meta-testing on each new task. The benchmark datasets used for evaluation and comparisons are Omniglot, MiniImageNet and CIFAR-100. All experiments were conducted on the Pytorch platform with assistance from the Torchmeta library \cite{deleu2019torchmeta} for standard implementation of Prototypical Networks.

The base model architecture comprises four convolutional blocks, with each block consisting of one 3x3 convolution layer, one BatchNormalization layer, followed by ReLU activation, and MaxPool2d layer. There are 64 input and 64 output hidden channels for each block, except 1 or 3 input channels for the first block corresponding to the number of RGB channels for the input image. The optimizer used to meta-train the encoder network is Adam, with a learning rate of 0.001 to match that of the original prototypical network implementation. However, one point of dissimilarity with the original implementation is that, unlike the original paper, where the model was meta-trained for tasks with a larger number of classes than during meta-testing whereas, in our experiments, the number of ways of the classification task is the same for both the meta training phase as well as the meta-testing. The code is available at \href{https://github.com/manasgg44/improved_proto.git}{https://github.com/manasgg44/improved\_proto.git}. \\

\begin{table*}[h!]
\caption{Classification accuracies for our approach in comparison with state-of-the-art meta-learning models on the Omniglot dataset}
\label{sample-table}
\vskip 0.15in
\begin{center}
\begin{small}
\begin{sc}
\begin{tabular}{lcccr}
\toprule
Approach & 5w1s & 5w5s & 20w1s & 20w5s\\
\midrule
  Prototypical Networks  & 98.39 $\pm$0.05  &  99.59 $\pm$0.02 & 95.49 $\pm$0.02 & 98.76 $\pm$0.03 \\
  MAML  & 98.46 $\pm$0.33  & 99.60$\pm$ 0.12 & 95.62 $\pm$ 0.31 & 98.69 $\pm$0.18 \\
  Matching Networks & 98.11 $\pm$0.11  & 98.9$\pm$ 0.20 & 93.8$\pm$ 0.21 & 98.5$\pm$ 0.11  \\
  Ours & \textbf{98.69}$\pm$ \textbf{0.04}  & \textbf{99.61}$\pm$ \textbf{0.01} & \textbf{95.67}$\pm$ \textbf{0.02} & \textbf{98.77}$\pm$ \textbf{0.03} \\
\bottomrule
\end{tabular}
\end{sc}
\end{small}
\end{center}
\vskip -0.1in
\end{table*}

\textbf{Omniglot} : The Omniglot dataset \cite{lake2019omniglot} includes 28x28 resolution images of 1623 distinct handwritten characters from 50 different alphabets (20 images of each character). Each character, regardless of language, is treated as a separate class in the classification task. The training and test classes for each run of our experiment are chosen at random from this pool of classes. The images pass through the four convolutional blocks of the encoder model and get flattened to a size 64-dimensional vector.

For the Omniglot dataset, the model augmentation consists of adding a linear layer consisting of n number of neurons (where n is the number of ways in the classification task) at the end of the base model. Moreover, during meta-testing, this dataset requires very few steps (in the range of 10 to 20) of optimization with the support set for optimal results that supersede the original Prototypical Networks both in terms of better mean accuracy and lower standard deviation. The results are shown in Table I. The optimal learning rate for the fine-tuning step is 1e-4 for all the layers of the augmented model. Another important observation in the results is that irrespective of the values of n and k in the n-way k shot setting, the accuracy of the proposed model reduces with an increment in the number of steps of cross-entropy optimization, a condition that can be attributed to overfitting.\\

\textbf{CIFAR-100} : This CIFAR-100 dataset \cite{krizhevsky2009learning} is identical to the CIFAR-10 with the exception that it comprises 100 classes with 600 photos per class. Out of the 600 images, 500 are training images and 100 are testing images. Moreover, the 100 classes are divided into 20 superclasses, and as such each image has a "fine" and a "coarse" designation, indicating the class and superclass respectively, to which it belongs. On passing the CIFAR-100 images through the same model architecture as described above, the resultant feature map is of size 2x2, which when flattened along 64 filters or channels, becomes 256 dimensions. This vector is then used for nearest centroid classification as is done in conventional prototypical networks.

For the CIFAR-100 dataset, the model is augmented in such a way that the feature map gets reduced to 1x1 before it is flattened for classification by a linear layer. In other words, the augmented model consists of an addition of another convolutional block at the end, followed by a linear layer. After fine-tuning, this kind of augmented model gives empirically better results than adding just a single linear classifier layer. The output channel width for the additional convolutional blocks are 100 and 25 for the 1-shot and 5-shot settings respectively. The number of steps of optimization required for the 1 shot setting and 5 shot setting for best results are 5 and 100. This shows that in the case of fewer shots, the encoder is more sensitive to changes in its parameters and hence a wider convolutional block and fewer steps of updates explain the reason for its optimal results, as shown in Table 2.\\

\begin{table}[t]
\caption{Classification accuracies for our approach in comparison with Prototypical Network on the CIFAR-100 dataset}
\label{sample-table}
\vskip 0.15in
\begin{center}
\begin{small}
\begin{sc}
\begin{tabular}{lcccr}
\toprule
Setting & Prototypical  & Our approach\\
\midrule
  5 way 1 shot  & 32.69 $\pm$ 0.22 & \textbf{32.80$\pm$ 0.26} \\
  5 way 5 shot  & 42.84 $\pm$ 0.16 & \textbf{44.01$\pm$ 0.18} \\
\bottomrule
\end{tabular}
\end{sc}
\end{small}
\end{center}
\vskip -0.1in
\end{table}

\textbf{MiniImageNet} : The MiniImageNet dataset\cite{vinyals2016matching} is a subset of the ImageNet dataset and consists of a total of 100 classes with 600 images per class. The training and testing split of the classes is done as 80/20. Each image is of 84x84 pixels and has 3 channels. As a result, after passing through the four convolutional blocks, the feature map size reduces to 5x5 and the size of the flattened feature vector becomes 1600 dimensions. 

For the MiniImagenet dataset, the augmented encoder consists of two additional blocks of convolutional layers, with output channels of 64, 64. The inner learning rate was set to 1e-7 for a count of 20 steps of optimization. The best results of the MiniImageNet dataset obtained are shown in Table 3. 

\begin{table}[t]
\caption{Classification accuracies for our approach in comparison with Prototypical Network on the MiniImageNet dataset}
\label{sample-table}
\vskip 0.15in
\begin{center}
\begin{small}
\begin{sc}
\begin{tabular}{lcccr}
\toprule
Setting & Prototypical  & Our approach\\
\midrule
  5 way 1 shot  & 46.44 $\pm$ 0.14 & \textbf{46.45$\pm$ 0.02} \\
  5 way 5 shot  & 63.22 $\pm$ 0.02 & \textbf{63.24$\pm$ 0.01} \\
\bottomrule
\end{tabular}
\end{sc}
\end{small}
\end{center}
\vskip -0.1in
\end{table}

\section{Conclusion}
\label{submission}
In this paper, we have proposed an intuitive method for improving upon the widely used Prototypical Networks algorithm based on the idea that training the model using the support set during meta-testing can enable the model to adapt to the change in the task and improve upon the accuracy of the model. For this purpose, the model was augmented with an additional classifier at the end and trained with cross-entropy loss using the task support set. The algorithm was tested on three benchmark data sets, and the results have confirmed that the proposed algorithm works better than most state-of-the-art meta-learning algorithms for classification.

\bibliographystyle{unsrtnat}
\bibliography{references}  






\end{document}